%% file: root.tex
\documentclass[letterpaper, 10 pt, conference]{ieeeconf}  

\IEEEoverridecommandlockouts                              

\usepackage{amsmath,amssymb,amsfonts}
\usepackage{mathtools}
\usepackage[dvipdfmx]{graphics}
\usepackage{cite}
\usepackage{algorithm}
\usepackage{algpseudocode}
\usepackage{array}
\usepackage[caption=false]{subfig}
\usepackage{stfloats}
\usepackage{url}

\usepackage{enumerate}
\usepackage{color}
\usepackage{xcolor}
\usepackage{subfiles}
\usepackage{comment}
\usepackage{bm}
\usepackage{bbm}

\newtheorem{remark}{Remark}

\title{\LARGE \bf
Communication-Free Distributed Multi-Robot Task Allocation under Partial Observations Using Labeled Multi-Bernoulli Filtering
}

\author{Takumi Ito$^{1}$ and Akiya Kamimura$^{1}$
\thanks{*This work was supported by JST K Program, Grant Number JPMJKP23G2.}
\thanks{$^{1}$Takumi Ito and Akiya Kamimura are with National Institute of Advanced Industrial Science and Technology (AIST), Tsukuba, Ibaraki 305-8568, Japan
        {\tt\small \{takumi.ito, kamimura.a\}@aist.go.jp}}%
}

\begin{document}

\maketitle
\thispagestyle{empty}
\pagestyle{empty}

\begin{abstract}
This paper proposes a communication-free multi-robot task allocation framework based solely on local observations. In this study, tasks are defined as reaching target locations.
Each robot estimates the positions of neighboring robots using a Labeled Multi-Bernoulli (LMB) filter and independently assigns tasks through a greedy auction-based strategy. 
By continuously updating state estimates and reallocating tasks during execution, the proposed method enables decentralized coordination without explicit communication. 
Monte Carlo simulations demonstrate that the proposed method enables effective cooperative task allocation without inter-robot communication while remaining robust to measurement clutter and observation uncertainty.
\end{abstract}

\subfile{text/introduction}
\subfile{text/modeling}
\subfile{text/method}
\subfile{text/simulation}


\section{Conclusion}

This paper presented a communication-free multi-robot task allocation framework that combined LMB filtering with auction-based task allocation. Each robot estimated neighboring robots from local observations and independently determined task assignments without communication.

Monte Carlo simulations under various clutter intensities showed that the LMB-based approach achieved higher task completion success rates and fewer task reassignments than KF- and PHD-based methods while maintaining competitive completion times and travel distances. These results demonstrate the importance of preserving robot identities and existence probabilities for communication-free task allocation under uncertain observations.

Future work includes establishing convergence guarantees for the proposed framework and improving performance in larger-scale scenarios. Another important direction is extending the proposed approach to more practical task settings.

\bibliographystyle{IEEEtran}
\bibliography{biblio}

\end{document}

%% file: text/introduction.tex
\section{Introduction}

Reliable cooperative decision-making is essential for multi-robot systems (MRSs) operating in challenging environments. Applications such as environmental monitoring \cite{funada_2024_coverage}, disaster response \cite{alotaibi_2019f_sar}, and transportation \cite{gao_2025_transportation} have been studied as MRS missions. In these applications, robots must decide how to distribute tasks among themselves based on available information. Consequently, Multi-Robot Task Allocation (MRTA) \cite{gerkey2004mrtataxonomy} has become a fundamental research topic and is expected to grow in importance as robotic platforms and tasks continue to diversify.

MRTA methods are classified into centralized \cite{ito2025multimode,gennaro2022resilient} and distributed approaches \cite{choi_consensus_based_2009}. Centralized methods compute globally optimal allocations using information collected from all robots. In contrast, distributed methods improve scalability and robustness by allowing each robot to make decisions independently. However, most distributed approaches rely on inter-robot communication to exchange information and coordinate actions. 

In many practical scenarios, communication may be unreliable or unavailable due to environmental constraints or limited communication ranges. 
As a result, enabling robots to perform cooperative task allocation without inter-robot communication remains an important challenge.
In this case, robots must rely solely on onboard sensing limited to nearby teammates. 
Furthermore, practical sensors are affected by limited sensing range, measurement noise, missed detections, and clutter. While several studies have investigated communication-free cooperation, many assume ideal observations or access to task-related information from neighboring robots \cite{mayya_closed_loop_2019, Dimos2016communicationfree}. 

This challenge is closely related to the problem of multi-target tracking. Recent advances in Random Finite Set (RFS)-based multi-target tracking have demonstrated strong robustness against uncertain observations. 
However, directly applying existing RFS-based estimators to MRTA is not straightforward. When tasks require a specific number of robots, robots must reason not only about where other robots are likely located but also about how many robots are available for each task. Density-based representations such as the PHD do not explicitly maintain unique target identities or individual existence probabilities. Consequently, they may lose information required for cardinality-aware task allocation decisions.

\begin{figure}[t]
    \centering
    \includegraphics[width=\linewidth, trim=15mm 67mm 9mm 150mm, clip]{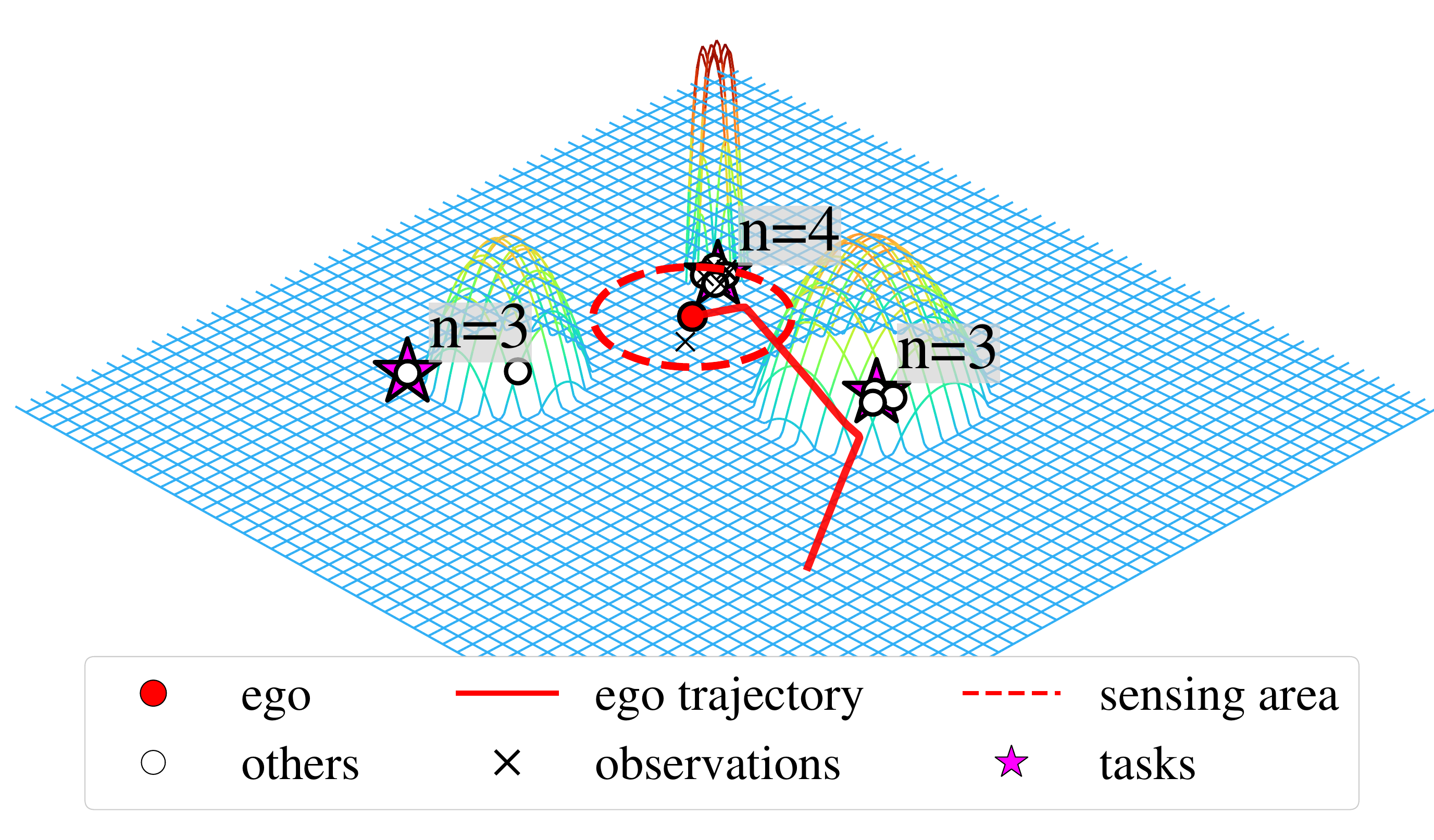}%
    \caption{Example scenario. Each robot decides its task based on partial observations of nearby robots. The value $n$ denotes the number of robots required to complete a task. The black crosses indicate observations that include both robot detections and clutter, and the background mesh represents the estimated locations of other robots.}
    \label{fig:teaser}
\end{figure}

As illustrated in Fig.~\ref{fig:teaser}, we consider a communication-free MRTA scenario in which robots with limited sensing ranges determine their tasks using current and past observations. Since observations are partial and subject to noise, missed detections, and clutter, robust estimation of neighboring robots is essential for reliable task allocation.
To address this challenge, we propose a communication-free MRTA framework based on the Labeled Multi-Bernoulli (LMB) filter \cite{reuter_lmb_2014}. The LMB filter maintains existence probabilities and labels for individual objects within the Random Finite Set framework. In this work, by modeling the detection probability according to the sensing distance, robots within the sensing range are updated with measurements, while those outside the sensing range are predicted without measurement updates. This behavior is naturally realized in the LMB framework, enabling communication-free task allocation under realistic sensing conditions.

The contributions of this paper are threefold:
1) We establish a novel connection between RFS-based multi-target tracking and communication-free MRTA.
2) We propose an LMB-based estimation and task-allocation framework that maintains robot identities under partial observations and explicitly accounts for task cardinality requirements. The incorporated sensing-aware detection probability model allows measurement updates for visible robots and prediction-only tracking for robots outside the sensing range.
3) We provide a systematic comparison of KF-, PHD-, and LMB-based estimation representations and their impact on task allocation performance, particularly in highly cluttered environments under realistic operational conditions, including collision avoidance among robots.
To the best of our knowledge, this is the first study to integrate the LMB filter with communication-free MRTA in a scenario where robots continuously estimate the global robot distribution from partial observations while navigating through the environment.

\subsection*{Related Work}

MRTA has been extensively studied in the multi-robot systems community and systematically classified by Gerkey and Matari\'c \cite{gerkey2004mrtataxonomy}. According to this taxonomy, the problem considered in this paper belongs to the ST-MR-IA category, where each robot executes a single task (ST), some tasks require multiple robots (MR), and task assignments are made instantaneously (IA). 

Several communication-free coordination methods have been proposed. Dai et al. \cite{Dai2019} predicted the actions of other robots using an incomplete-information game. Mayya et al. \cite{mayya_closed_loop_2019} proposed a task allocation strategy based on the observed positions and task assignments of neighboring robots. Guo et al. \cite{Dimos2016communicationfree} developed a cooperative control method that maintains inter-robot distance constraints without communication. However, these approaches generally assume reliable observations and do not explicitly address uncertainty arising from noise, missed detections, and clutter.

Random Finite Set (RFS) based filters, including the PHD \cite{vo_gmphd_2026} and LMB \cite{reuter_lmb_2014} filters, have demonstrated strong robustness in multi-target tracking under noisy observations and clutter. RFS-based estimation has also been applied to robotic problems \cite{papaioannou_2021_cooperative,papaioannou_2020_jointly,dames_distributed_2020}. Nevertheless, their application to MRTA remains largely unexplored.

%% file: text/modeling.tex
\section{Problem Formulation}
\label{sec:problem_formulation}

We consider a team of robots that cooperatively performs spatially distributed tasks, where completing each task requires a prescribed number of robots to reach a target, without direct inter-robot communication. The task locations are assumed to be known to all robots a priori. Each robot knows its own position in a global coordinate frame and observes nearby robots using a range-limited onboard sensor. The measurements are uncertain, may contain clutter, and do not include robot identifiers. Consequently, each robot must determine its task assignment from an unlabeled finite set of local measurements.

\subsection{Robot and Task Models}

Let $\mathcal{R}=\{1,\ldots,N_r\}$ denote the index set of robots operating in a bounded planar workspace $\mathcal{W}\subset\mathbb{R}^{2}$. The position and velocity of robot $i\in\mathcal{R}$ are denoted by $\bm p_i\in\mathcal{W}$ and $\bm v_i\in\mathbb{R}^{2}$, respectively, and satisfy
\begin{subequations}
    \label{eq:robot_dynamics}
    \begin{align}
        \dot{\bm p}_i &= \bm v_i, \\
        \dot{\bm v}_i &= \mu\left(\bm u_{i,\mathrm{nom}}-\bm v_i\right),
        \label{eq:input_dynamics}
    \end{align}
\end{subequations}
where $\bm u_{i,\mathrm{nom}}\in\mathbb{R}^{2}$ is the commanded velocity and $\mu>0$ is the velocity-response coefficient.

Let $\mathcal{T}=\{1,\ldots,N_t\}$ denote the index set of tasks. To simplify the problem formulation, task $t\in\mathcal{T}$ is modeled as a location-visiting task specified by its target position $\bm q_t \in \mathcal{W}$ and demand $n_t \in \mathbb{N}_{>0}$, where $n_t$ denotes the number of robots required for completion, and we assume that $\sum_{t=1}^{N_t} n_t \leq N_r$. A task is regarded as completed when the required number of robots are within a specified distance of the target position.
All the task information $\bm q_t,\,n_t,\,t\in\mathcal{T}$ is known to every robot before task allocation begins.

Let $a_i\in\mathcal{T}$ denote the task selected by robot $i$. The objective is to determine task assignments using only locally available information such that
\begin{equation}
    \sum_{i\in\mathcal{R}}\mathbb{I}[a_i=t]=n_t,
    \qquad \forall t\in\mathcal{T},
    \label{eq:task_number_satisfaction}
\end{equation}
where $\mathbb{I}[\cdot]$ denotes the indicator function, which equals $1$ if its argument is true and $0$ otherwise, and all robots assigned to task $t$ reach a neighborhood of the target position $\bm q_t$, i.e.,
\begin{equation}
    \|\bm p_i-\bm q_t\| \leq d_\mathrm{comp},
    \quad
    \forall i\in\mathcal{R}\ \text{such that}\ a_i=t,
    \label{eq:task_distance_satisfaction}
\end{equation}
where $d_\mathrm{comp}>0$ is a prescribed distance threshold.

\subsection{Observation Model}

Each robot is equipped with a range-limited onboard sensor that observes neighboring robots. Let
\begin{equation}
    \mathcal S_i =
    \left\{
        \bm z \in \mathbb R^2 : \|\bm z-\bm p_i\| \le d_{\mathrm{s1}}
    \right\}
\end{equation}
denote the sensing region of robot $i$, and let
\begin{equation}
    d_{ii'} = \|\bm p_{i'}-\bm p_{i}\|
\end{equation}
denote the distance between robots $i$ and $i'$.
The detection probability is modeled as
\begin{equation}
    p_{D,ii'}(d_{ii'}) \!=\!\!
    \begin{cases}
        p_D^{\max}, & \!\!d_{ii'} \!\le\! d_{\mathrm{s0}}, \\[1mm]
        p_D^{\max}\!\!
        \left(
            \!1\!-\!\dfrac{d_{ii'}\!-\!d_{\mathrm{s0}}}{d_{\mathrm{s1}}\!-\!d_{\mathrm{s0}}}
        \right),
        & \!\!d_{\mathrm{s0}} \!<\! d_{ii'} \!<\! d_{\mathrm{s1}}, \\[2mm]
        0, & \!\!d_{ii'} \!\ge\! d_{\mathrm{s1}}.
    \end{cases}
    \label{eq:detection_probability}
\end{equation}
Conditioned on successful detection, the position measurement follows
\begin{align}
    \label{eq:measurement_model}
    \bm z_{ii'} &= \bm p_{i'} + \bm w_{ii'},\\
    \bm w_{ii'} &\sim \mathcal N
    \left(
        \bm 0, \sigma_{ii'}^2 \bm I
    \right),\notag\\
    \sigma_{ii'} &= \sigma_0 + a_\sigma d_{ii'},\notag
\end{align}
where $\mathcal{N}(\bm m,\bm P)$ is a Gaussian distribution with mean $\bm m$ and covariance $\bm P$, and $a_\sigma > 0$ is a distance-dependent uncertainty scaling parameter.

Given the detection and measurement models in \eqref{eq:detection_probability} and \eqref{eq:measurement_model}, we define the observations at time step $k$ as follows.
Let $\delta_{ii',k}\sim\mathrm{Bernoulli}\!\left(p_{D,ii'}(d_{ii',k})\right)$ denote the detection event, where $\delta_{ii',k}=1$ with probability $p_{D,ii'}(d_{ii',k})$ and $\delta_{ii',k}=0$ otherwise. The set of target-originated measurements is
\begin{equation}
    \Theta_k^{(i)} =
    \left\{
        \bm z_{ii',k} : i' \in \mathcal R \setminus \{i\},~\delta_{ii',k}=1
    \right\}.
\end{equation}
In addition, clutter measurements are present. The number of clutter measurements is modeled as $N_{c,k}^{(i)} \sim \mathrm{Poisson}(\lambda_c)$, where $\lambda_c$ represents the expected number of clutter measurements per scan. Each clutter measurement is independently sampled from a uniform distribution over the sensing region $\mathrm{Unif}(\mathcal S_i)$, yielding the clutter measurement set
\begin{equation}
    K_k^{(i)} =
    \left\{
        \bm c_{\ell,k}^{(i)}
    \right\}_{\ell=1}^{N_{c,k}^{(i)}},
    \qquad
    \bm c_{\ell,k}^{(i)}
    \overset{\mathrm{i.i.d.}}{\sim}
    \mathrm{Unif}(\mathcal S_i).
    \label{eq:poission_clutter}
\end{equation}
The complete measurement set available to robot $i$ is
\begin{equation}
    Z_k^{(i)} = \Theta_k^{(i)} \cup K_k^{(i)}.
\end{equation}
The measurement set does not contain unique identifiers; each robot observes only an unlabeled finite set of position measurements. Based on the measurement history $Z_{1:k}^{(i)}$, each robot recursively estimates the states of both observed and unobserved neighboring robots via Bayesian filtering and determines its task assignment without inter-robot communication.

%% file: text/method.tex
\section{LMB Filter Based Task Allocation}
\label{sec:method}

\subsection{Labeled Multi-Bernoulli State Estimation}
\label{sec:lmb_estimation}

The proposed method uses a labeled multi-Bernoulli (LMB) representation to estimate the states of other robots. By maintaining independent Bernoulli components for each robot, it can distinguish individual robots and estimate their states while handling data association under missed detections and clutter.
Each robot estimates the states of other robots from its unlabeled local measurement set $Z_k^{(i)}=\{\bm z_{1,k},\ldots,\bm z_{m_k,k}\}$, where $m_k=|Z_k^{(i)}|$, using a labeled multi-Bernoulli (LMB) filter \cite{reuter_lmb_2014}. Hereafter, the superscript $(i)$ indicating the ego robot is omitted.

Let the object state at time step $k$ be $\bm x_k=[\bm p_{k}^\top~\bm v_{k}^\top]^\top$. 
The linear Gaussian transition and measurement models are defined as
\begin{subequations}
\begin{align}
    \bm x_k &= \bm F\bm x_{k-1}+\bm \eta_{k-1}, & \bm \eta_{k-1}&\sim\mathcal{N}(\bm0,\bm Q), \\
    \bm z_{k} &= \bm H\bm x_k+\bm \epsilon_{k}, & \bm \epsilon&\sim\mathcal{N}(\bm0,\bm  R_{k}).
    \label{eq:lmb_models}
\end{align}
\end{subequations}

The states of the other robots are represented by an LMB random finite set. Each Bernoulli component is characterized by an existence probability $r_k^{(\ell)}$ and a spatial density $p_k^{(\ell)}(\bm x)$.
Each component is associated with a unique label $\ell=(\tau_b,\iota_b)\in\mathcal{L}_k$, where $\mathcal{L}_k$ is a label set, $\tau_b$ denotes the birth time, and $\iota_b$ is a unique index used to distinguish components born at the same time.
To reduce computational complexity, we use an approximate single-Gaussian LMB implementation and retain only one Gaussian component for each Bernoulli density. In this work, the spatial density of each component is modeled as a Gaussian distribution:
\begin{equation}
    p_{k}^{(\ell)}(\bm x)=
    \mathcal{N}\!\left(\bm x;\bm m_{k}^{(\ell)},
    \bm P_{k}^{(\ell)}\right).
    \label{eq:lmb_gm}
\end{equation}
where $\bm m_{k}^{(\ell)}$ and $\bm P_{k}^{(\ell)}$ denote the mean vector and covariance matrix, respectively.

\subsubsection{Prediction}
As in standard Bayesian filtering, the first step of the LMB filter is prediction.
For an existing component, the predicted existence probability and Gaussian parameters are
\begin{subequations}
\begin{align}
    r_{k|k-1}^{(\ell)} &=p_S r_{k-1}^{(\ell)}, \\
    \bm m_{k|k-1}^{(\ell)} &=\bm F\bm m_{k-1}^{(\ell)}, \\
    \bm P_{k|k-1}^{(\ell)} &=\bm F\bm P_{k-1}^{(\ell)}\bm F^\top+\bm Q, \\
    &\qquad \ell\in\mathcal L_{k-1}.\notag
    \label{eq:lmb_prediction}
\end{align}
\end{subequations}
where $p_S$ is the survival probability. 
Assuming that robots rarely break down or disappear, we set $p_S=1-10^{-6}$ in the implementation.

\subsubsection{Update and data association}
The next step updates the predicted density using the measurement set $Z_k$.
Because the measurements generated by other robots and clutter are unlabeled, the filter must determine which measurement corresponds to each predicted Bernoulli component. This data association is represented by a mapping from the label space $\mathcal{L}_{k-1}$ to the augmented measurement set $\mathcal{Z}_k := \{1,\ldots,m_k\} \cup \{\bot\}$, where $\bot$ denotes a missed detection.

For measurement $j=1,\ldots,m_k$, the measurement covariance is denoted by $\bm R_{j,k}=\sigma_{ij,k}^{2}\bm I$, as defined in \eqref{eq:measurement_model}.
For measurement $j=1,\ldots,m_k$ and Bernoulli component $\ell \in \mathcal L_{k-1}$, the innovation covariance and Kalman update are
\begin{subequations}
\label{eq:lmb_kalman_update}
\begin{align}
    \bm S_{j,k}^{(\ell)} &= \bm H\bm P_{k|k-1}^{(\ell)}\bm H^\top+\bm R_{j,k}, \\
    \bm K_{j,k}^{(\ell)} &= \bm P_{k|k-1}^{(\ell)}\bm H^\top \big(\bm S_{j,k}^{(\ell)}\big)^{-1}, \\
    \bm m_{j,k|k}^{(\ell)} &= \bm m_{k|k-1}^{(\ell)} +\bm K_{j,k}^{(\ell)} \left(\bm z_{j,k}-\bm H\bm m_{k|k-1}^{(\ell)}\right), \\
    \bm P_{j,k|k}^{(\ell)} &= (\bm I-\bm K_{j,k}^{(\ell)}\bm H) \bm P_{k|k-1}^{(\ell)},\\
    &\qquad \forall \bm z_{j,k}\in Z_k,~\forall \ell \in \mathcal L_{k-1}.\notag
\end{align}
\end{subequations}
The posterior existence probability is updated as,
\begin{align}
    r_{j,k|k}^{(\ell)} =
    \begin{cases}
    1, & z_j \neq \bot,\\
    \frac{(1-p_D)r_{k|k-1}^{(\ell)}}{1-p_D r_{k|k-1}^{(\ell)}}, & z_j = \bot.
    \end{cases}
    \label{eq:probability_update}
\end{align}
For a predicted component $\ell$, its detection probability $p_D^{(\ell)}$ follows \eqref{eq:detection_probability}.
The likelihood ratio for assigning $\bm z_{j,k}$ to label $\ell$ is
\begin{equation}
    L_{\ell,j}=
    \frac{r_{k|k-1}^{(\ell)}p_D^{(\ell)}}{\kappa}
    \mathcal{N}\!\left(
    \bm z_{j,k};
    \bm H\bm m_{k|k-1}^{(\ell)},
    \bm S_{j,k}^{(\ell)}
    \right),
    \label{eq:lmb_likelihood_ratio}
\end{equation}
where $\kappa=\lambda_c/(\pi d_{\mathrm{s1}}^2)$ is the clutter intensity under the clutter model in \eqref{eq:poission_clutter}.
The likelihood ratio for a missed detection is
\begin{equation}
    L_{\ell,\bot} = 1-r_{k|k-1}^{(\ell)}p_D^{(\ell)}.
    \label{eq:missed_likelihood}
\end{equation}

Let $\theta:\mathcal{L}\to\mathcal Z_k$ denote a label-to-observation assignment hypothesis.
Enumerating all valid measurement-to-label associations is computationally expensive. 
Therefore, we use Murty's algorithm \cite{murty_1068_ranking} to generate the $K_{\mathrm{A}}$ highest-scoring association hypotheses $\theta_1,\dots,\theta_{K_A}$, which minimize the cost
\begin{align}
    J(\theta) = \sum_{\ell\in\mathcal L_{k|k-1}} -\log L_{\ell,\theta(\ell)}.
\end{align}
The weight of assigning label $\ell$ to event $j\in\mathcal Z_k$ is obtained by marginalizing the association hypotheses,
\begin{align}
    w_{\ell,j} = \sum_{\theta_a(\ell)=j} \left(\prod_{\ell'\in\mathcal L_{k|k-1}} L_{\ell',\theta_a(\ell')}\right).
\end{align}

Finally, the posterior existence probability is calculated as
\begin{align}
    r_{k|k}^{(\ell)} = 
    \frac{\sum_{j\in\mathcal{Z}_k} w_{\ell,j} r_{j,k|k}^{(\ell)}}{\sum_{j\in\mathcal{Z}_k} w_{\ell,j}}
    \label{eq:r_update_using_hypothesis}
\end{align}
For the posterior Gaussian density, only the highest-weight component is retained.
Therefore, the posterior spatial density is
\begin{align}
    p_{k|k}^{(\ell)}(\bm x) = \mathcal N\!\left(\bm x; \bm m_{j^\star,k|k}^{(\ell)}, \bm P_{j^\star,k|k}^{(\ell)} \right),
\end{align}
where
\begin{align}
    j^\star = \arg\max_{j\in\mathcal Z_k} w_{\ell,j}.
\end{align}

\begin{remark}
When a Bernoulli component lies outside the sensing range, its detection probability becomes zero, i.e., $p_D^{(\ell)}=0$. In this case, the association likelihood in \eqref{eq:lmb_likelihood_ratio} vanishes, while the missed-detection likelihood in \eqref{eq:missed_likelihood} remains. Consequently, if $\theta(\ell)=\bot$, substituting $p_D^{(\ell)}=0$ into \eqref{eq:probability_update} and \eqref{eq:r_update_using_hypothesis} yields $r_{k|k}^{(\ell)}=r_{k|k-1}^{(\ell)}$, i.e., the existence probability remains unchanged. The track is propagated only through prediction. In contrast, tracks within the sensing range are updated by measurements, and false tracks are gradually removed as their existence probabilities decrease through \eqref{eq:probability_update} when no consistent observations are received. Therefore, previously observed robots outside the current field of view remain available for task-allocation and collision-avoidance decisions.
\end{remark}

\subsubsection{Track Birth and Pruning} \label{ssec:birth_pruning}
The update described above only associates the current measurements with existing tracks. Therefore, it cannot directly represent a robot that has newly entered the sensing region. To handle such robots, measurements that cannot be explained by the existing tracks are used to initialize new Bernoulli components. These components are introduced into the filtering process at the next time step.
A measurement is considered unexplained if it lies outside the two-dimensional chi-square gate of every existing track. Specifically, measurement $\bm z_{j,k}$ is used to initialize a birth component if
\begin{align*}
    \left(\bm z_{j,k}-\bm H\bm m_{k|k-1}^{(\ell)}\right)^\top
    \left(\bm S_{j,k}^{(\ell)}\right)^{-1}
    \left(\bm z_{j,k}-\bm H\bm m_{k|k-1}^{(\ell)}\right)
    >
    \gamma_{\mathrm B}
\end{align*}
for all $\ell\in\mathcal L_k$, where $\gamma_{\mathrm B}=5.99$ in the implementation, corresponding to a 95\% confidence region for a two-dimensional measurement.
For each unexplained measurement $\bm z_{j,k}$, a new Bernoulli component is initialized with $\bm m=[\bm z_{j,k}^\top~0~0]^\top,~r=0.01$. Its initial covariance is set to $\bm P_0$. We set $\mathrm{diag}(0.18,0.18,4.0, 4.0)$ in the implementation.

In a cluttered environment, repeatedly generating birth components can increase the number of tracks and the computational cost. We therefore prune components with low existence probabilities or excessive state uncertainty.
Specifically, components with $r^{(\ell)}<0.5$ or $\mathrm{tr}(P)>60$ are discarded.

Finally, the remaining Bernoulli components with existence probabilities satisfying $r \geq 0.5$ are ranked according to the uncertainty of their state estimates, measured by $\mathrm{tr}(\bm P)$. The $\sum_{t=1}^{N_t} n_t-1$ components with the smallest $\mathrm{tr}(\bm P)$ values are then provided to the task-allocation and collision-avoidance modules. This guarantees that at least one task remains available for the ego robot in the subsequent task-allocation algorithm.

\subsection{Greedy Auction-Based Task Assignment}
\label{subsec:greedy_auction_assignment}

The next step is auction-based task assignment. In this approach, each robot computes a bid for every task, and tasks are assigned based on the resulting bids. Using the estimated positions of other robots and the known task locations, each robot independently performs a greedy auction-based task assignment. Although the auction is executed locally, all robots use the same deterministic assignment rule.

For robot $i$ and task $j$, the distance between the robot and the task is defined as
\begin{equation}
    d_{ij} =
    \left\|
    \bm{p}_i - \bm{q}_j
    \right\|_2,
    \label{eq:distance_robot_task}
\end{equation}
where $\bm{p}_i$ is the position of robot $i$ and $\bm{q}_j$ is the position of task $j$. Similarly, for a tracked robot with label $\ell$ whose state estimate has mean $\bm m^{(\ell)}$, the distance to task $j$ is given by
\begin{equation}
    d_{\ell j} =
    \left\|
    \hat{\bm p}^{(\ell)} - \bm q_j
    \right\|_2,
    \label{eq:distance_other_robot_task}
\end{equation}
where $\hat{\bm p}^{(\ell)} = [\bm I_2~\bm O_2]\bm m^{(\ell)}$.
The bid value is computed by applying a softmax function to the distances:
\begin{equation}
b_{ij}
=
\frac{\exp(-d_{ij})}
{\sum_{l=1}^{N_t}\exp(-d_{il})}.
\label{eq:softmax_bid}
\end{equation}
Therefore, for each robot, the bids satisfy $\sum_{j=1}^{N_t} b_{ij}=1$.
The bid value of another robot, denoted by $b_{\ell j}$, is computed in the same manner.

Assignments are determined greedily in descending order of bids. Specifically, all robot--task bids $b_{ij}$ and $b_{\ell j}$ are sorted by their values and processed sequentially. A robot--task pair is accepted only when the robot has not yet been assigned to any task and the corresponding task has not reached its required number of robots. The process continues until a task is assigned to the ego robot. Since only $\sum_{t=1}^{N_t} n_t-1$ robots are provided from the estimation module, at least one task is always assigned to the ego robot.

To avoid frequent task switching, the assigned task is determined once every 20 time steps based on the majority vote of the assignments. Robots continue toward their current tasks unless the tasks become occupied by robots with higher bids, in which case they switch to alternative tasks. Repeating this process enables robots to distribute themselves among tasks without communication.

\subsection{Task Execution and Collision Avoidance Control}
\label{ssec:control_law}

This subsection provides the control law that enables the robots to execute their assigned tasks while avoiding collisions.
Let $a_i$ denote the task assigned to robot $i$. The nominal velocity command is given by
\begin{equation}
    \bm{u}_{i,\mathrm{nom}} = \mathrm{sat} \left(k_p \left(\bm{q}_{a_i} - \bm{p}_i \right), v_{\max} \right),
    \label{eq:proportional_controller}
\end{equation}
where $k_p$ is the control gain and $\mathrm{sat}(\cdot)$ limits the velocity
magnitude by $v_{\max}$.

To avoid collisions between robots, we utilize the Control Barrier Function (CBF) \cite{ames2019CBF}. The CBF provides a formal framework for ensuring that the system state remains within a safe set by enforcing safety constraints.
Let the safe distance between robots be $d_{\mathrm{safe}}$. For robot $i$, the safety constraint with respect to another robot is defined as
\begin{equation}
    h_{i\ell}(\bm{p}_i) = \|\bm{p}_i-\hat{\bm p}^{(\ell)}\|^2 - d_{\mathrm{safe}}^2,
\end{equation}
where $\hat{\bm p}^{(\ell)}$ is the estimated position of a neighboring robot by LMB filtering. The condition $h_{i\ell}\ge0$ indicates that the state belongs to the \textit{safe} (collision-free) set.
The robot dynamics \eqref{eq:robot_dynamics} can be rewritten into the control-affine form
\begin{equation}
    \dot{\bm x_i} = \bm f(\bm x_i) + \bm g(\bm x_i)\bm u_\mathrm{nom},
    \label{eq:control_affine}
\end{equation}
where $\bm x_i=[\bm p_i^\top~\bm v_i^\top]^\top$.
Following the high order CBF formulation, the safety constraint is expressed as
\begin{align}
    L_gL_f h_{i\ell}(\bm{x}_i)\,\bm{u}_\mathrm{nom} + L_f^2 h_{i\ell}(\bm{x}_i) + \gamma L_f h_{i\ell}(\bm{x}_i) & \notag\\
     + \gamma_2 \left(L_f h_{i\ell}(\bm{x}_i) + \gamma h_{i\ell}(\bm{x}_i) \right) &\ge 0, 
    \label{eq:cbf_constraint}
\end{align}
where $L_{\{f,g\}}$ denotes the Lie derivative with respect to the vector fields $\bm f$ and $\bm g$ in \eqref{eq:control_affine}, and $\gamma>0$ and $\gamma_2>0$ are class-$\mathcal{K}$ function coefficients. See \cite{ames2019CBF} for details.

At each control step, the collision-free control input is obtained by solving the quadratic program
\begin{align}
    \bm{u}^{*}_i = \arg\min_ {\bm{u}_i} &~ \frac{1}{2} \left\| \bm{u}_i - \bm{u}_{i,\mathrm{nom}} \right\|^2 \\
    \mathrm{subject~to} &~ \eqref{eq:cbf_constraint},~\forall \ell\in \mathcal{L}.\notag
\end{align}
As a result, the collision avoidance module minimally modifies the nominal command while enforcing collision-avoidance constraints with respect to the estimated neighboring-robot states.

By repeatedly executing the estimation, task-assignment, and control steps, the robots adapt their task assignments according to the locally estimated robot distribution while satisfying the required number of robots for each task.

%% file: text/simulation.tex
\section{Simulation}

Finally, simulations are conducted to evaluate the proposed method. The task-assignment performance is assessed through Monte Carlo simulations with randomly generated robot and task locations. As benchmarks, we consider a Kalman filter (KF) based on heuristic association and the Probability Hypothesis Density (PHD) filter, a widely used multi-target tracking method alongside the LMB filter.

\subsection{Comparison methods}

For comparison, only the robot-state estimation module in Section~\ref{sec:lmb_estimation} is replaced by the following methods, while the task allocation and control modules remain unchanged.

\subsubsection{Ground Truth (GT)}

As an ideal baseline, we consider a ground-truth setting in which the exact positions of all robots are available. Since all robots share identical state information, the task-allocation process yields identical assignment results across robots. This baseline represents an approximate upper bound on performance achievable with perfect robot-state information.

\subsubsection{Kalman Filter (KF)}

We employ a nearest-neighbor Kalman filter (KF) to estimate neighboring robot states. At each time step, all existing tracks are first propagated using the KF prediction step. The predicted tracks are then associated with newly received observations through nearest-neighbor data association based on Euclidean distance gating with $1.0$ m. Associated tracks are updated using the standard KF correction step, while unmatched observations initialize new tentative tracks.

To suppress false tracks caused by clutter, tentative tracks are confirmed only after three successful associations. Tracks that are repeatedly missed for five consecutive time steps while their predicted positions remain within the sensing range, or whose state uncertainty exceeds a threshold $\mathrm{tr}(P) > 60$, are removed. The resulting confirmed tracks are used for task allocation. Unlike the LMB filter, the extracted states do not maintain unique labels or individual existence probabilities.

\subsubsection{PHD Filter (PHD)}

We employ a PHD filter to estimate neighboring robot states. In particular, we use the Gaussian Mixture PHD (GM-PHD) filter \cite{vo_gmphd_2026}, a practical implementation of the PHD filter. The PHD filter represents the multi-target state as an intensity function modeled by a Gaussian mixture and does not require explicit data association between tracks and observations.

Target states are extracted from peaks of the Gaussian-mixture intensity. Specifically, Gaussian components whose weights exceed $w=0.5$ are regarded as robot-state estimates. Components whose uncertainty exceeds $\mathrm{tr}(P)>60$ are discarded. The resulting extracted states are used for task allocation.

\subsection{Simulation Setup}

\subsubsection{Setting}

In this work, we set $K_{\mathrm A}=20$, $p_S=1-10^{-6}$, $p_D^{\max}=0.9$, $\sigma_0=0.04$, $a_\sigma=0.04$, and $d_{\mathrm{safe}}=1~\mathrm{m}$. At a sensing distance of $1~\mathrm{m}$, the observation covariance is $0.08^2 I_3$. Under this setting, the probability that the observation error exceeds $d_{\mathrm{safe}}$ is approximately $10^{-32}$.


The simulation environment is a $50\,\mathrm{m}\times50\,\mathrm{m}$ square area, and each robot is equipped with a range-limited sensor with a sensing range of $d_{\mathrm{s0}}=4.5,\,d_{\mathrm{s1}}=5.5\,\mathrm{m}$. Two problem sizes are considered: $10$ robots and $3$ tasks, and $15$ robots and $4$ tasks. The initial robot positions are randomly sampled within the central $30\,\mathrm{m}\times30\,\mathrm{m}$ region of the environment, and task locations are generated randomly in the same manner.

A total of $200$ random initial states are generated. For each trial, task-allocation simulations based on KF-, PHD-, and LMB-based neighboring robot estimation are performed at clutter intensities of $\lambda_c\in\{0.0, 0.5, 1.0, 2.0\}$. The same set of $200$ random seeds is used for all estimation methods, ensuring that all methods are evaluated under identical scenarios.

The simulator and controller were implemented in ROS~2 and executed on a PC equipped with an Intel\textsuperscript{\textregistered} Core\texttrademark~Ultra~9~285K processor (24 cores).
The LMB filter used in this study was implemented in Python, based on Robertson's publicly available implementation \cite{robertson2022mlmb, robertsongithub}.

\subsubsection{Evaluation Metrics}

Since the proposed method continuously assigns tasks, a scenario is regarded as successful if the task satisfaction conditions in \eqref{eq:task_number_satisfaction} and \eqref{eq:task_distance_satisfaction} are satisfied within the 180~s timeout limit. To account for collision avoidance, the completion distance threshold $d_\mathrm{comp}$ was set to 1.5, 1.5, 2.0, 2.2, 2.5, 2.8, 3.0, and 3.3~m for task capacities 1 to 8, respectively.

The proposed framework estimates neighboring robots after they have once been observed and continues predicting their states even after they leave the sensing range. In contrast, robots that have never entered the sensing range cannot be estimated. Furthermore, efficient task allocation may reduce unnecessary robot motion and consequently decrease opportunities for observing neighboring robots. For these reasons, conventional multi-target tracking metrics such as OSPA are not well suited for evaluating the overall effectiveness of the proposed approach.
Instead, we evaluate the performance of the target application itself, namely task allocation, using the following metrics:
\begin{itemize}
    \item \textbf{Success rate}: The fraction of trials in which all tasks are completed with the required number of robots.
    
    \item \textbf{Travel distance}: The sum of travel distances of all robots in successful trials.
    
    \item \textbf{Completion time}: The time required to complete all tasks in successful trials.
    
    \item \textbf{Task reassignment}: The number of times robots change their assigned task in successful trials.
\end{itemize}
These metrics reflect the impact of estimation uncertainty on task allocation. Inaccurate estimates may cause inefficient assignments, leading to more task reassignments, longer travel distances, and increased completion times.

\subsection{Results and Discussion}

\begin{figure*}[t]
    \centering
    \subfloat[]{\includegraphics[height=4.2cm]{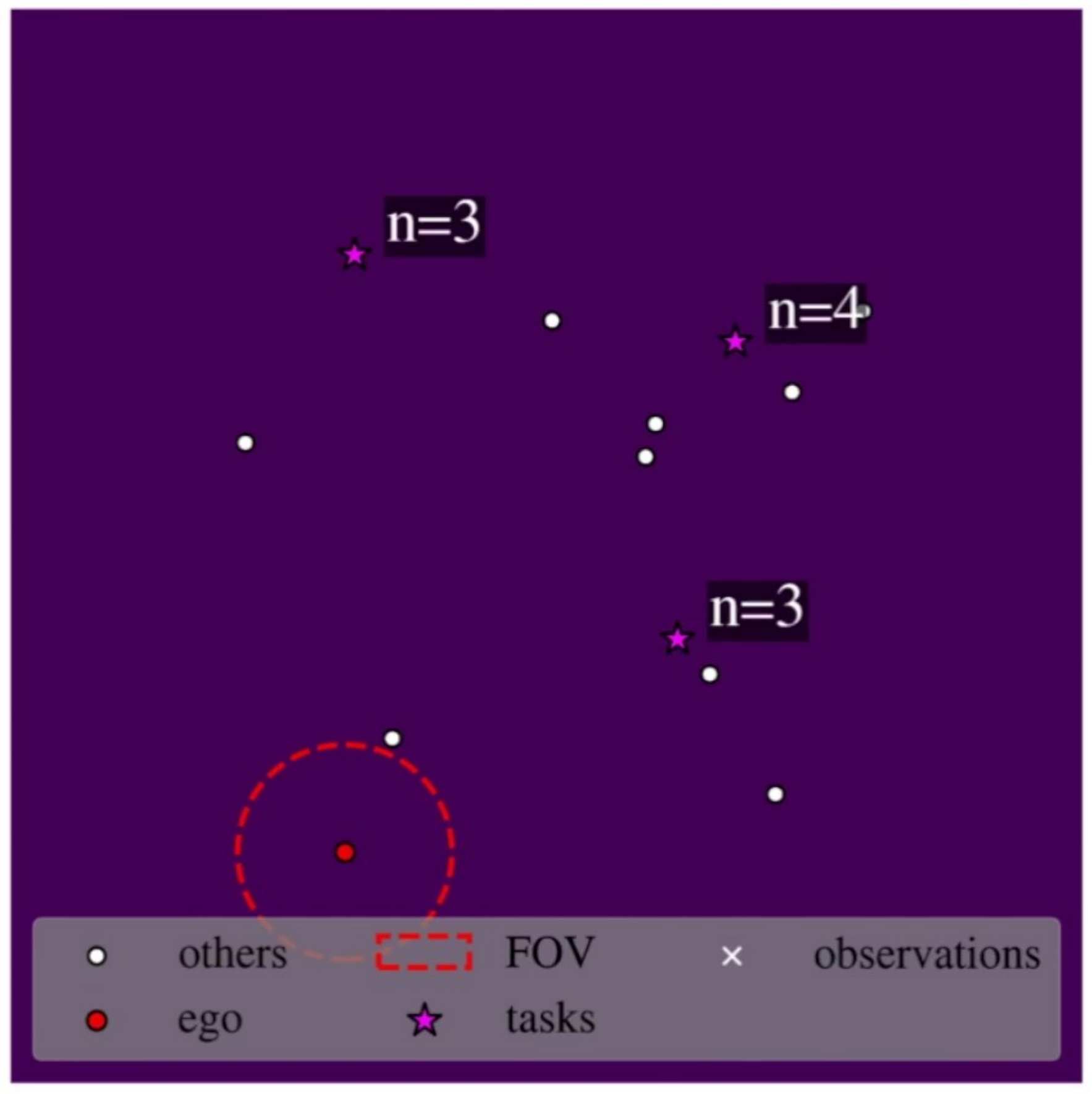}%
    \label{fig:sim_example0}}
    \hfil
    \centering
    \subfloat[]{\includegraphics[height=4.2cm]{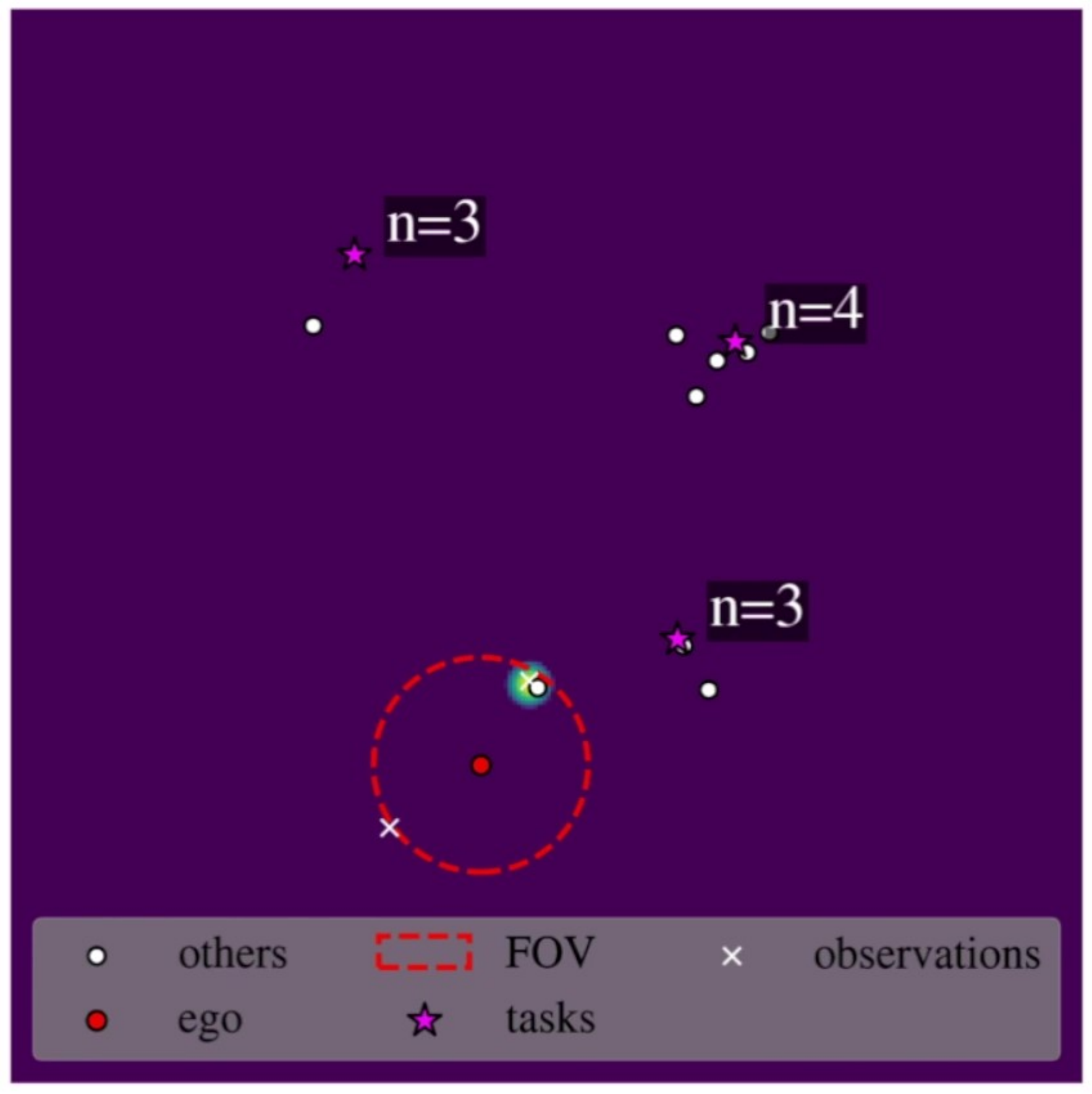}%
    \label{fig:sim_example8}}
    \hfil
    \centering
    \subfloat[]{\includegraphics[height=4.2cm]{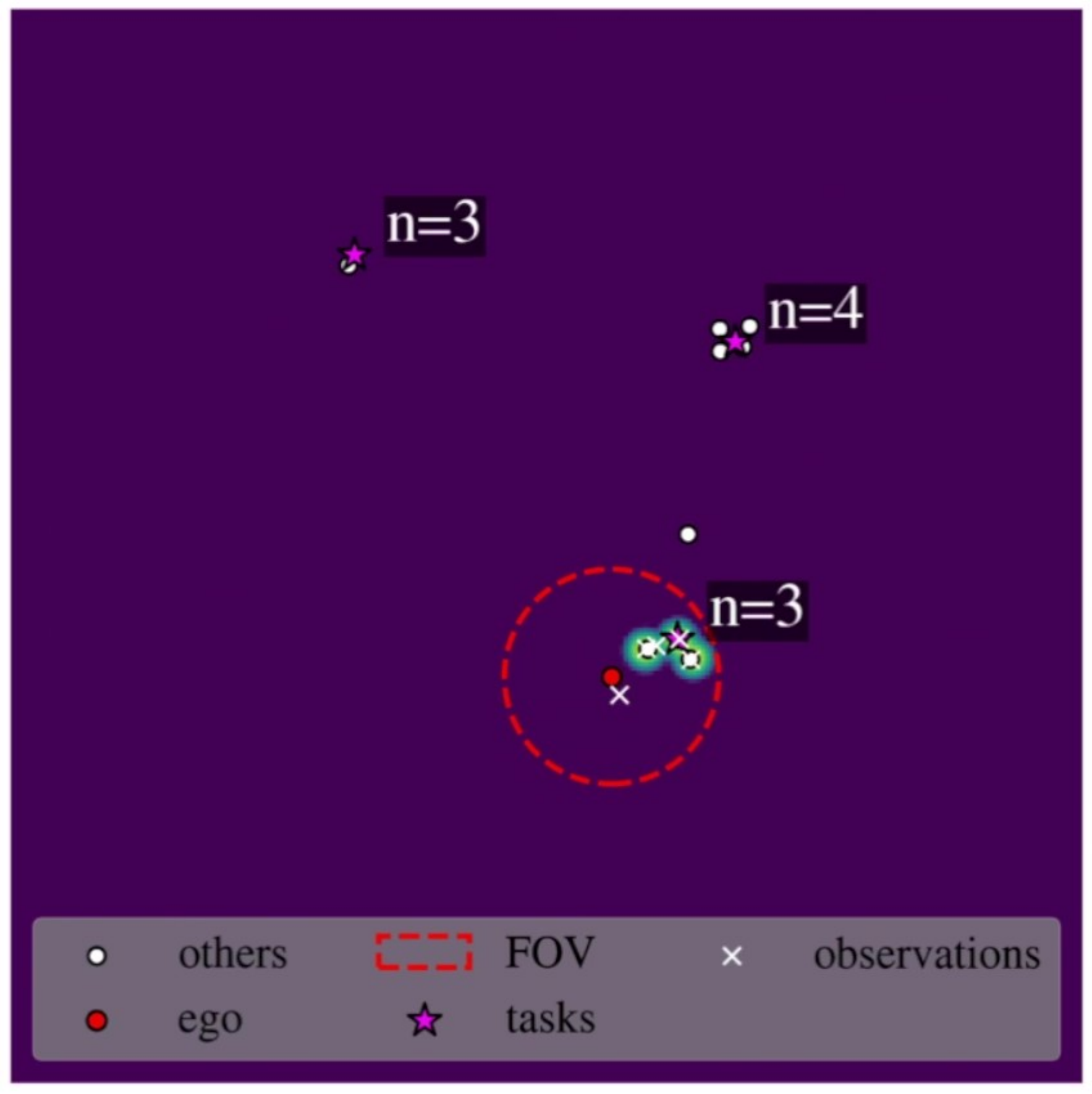}%
    \label{fig:sim_example18}}
    \hfil
    \centering
    \subfloat[]{\includegraphics[height=4.2cm]{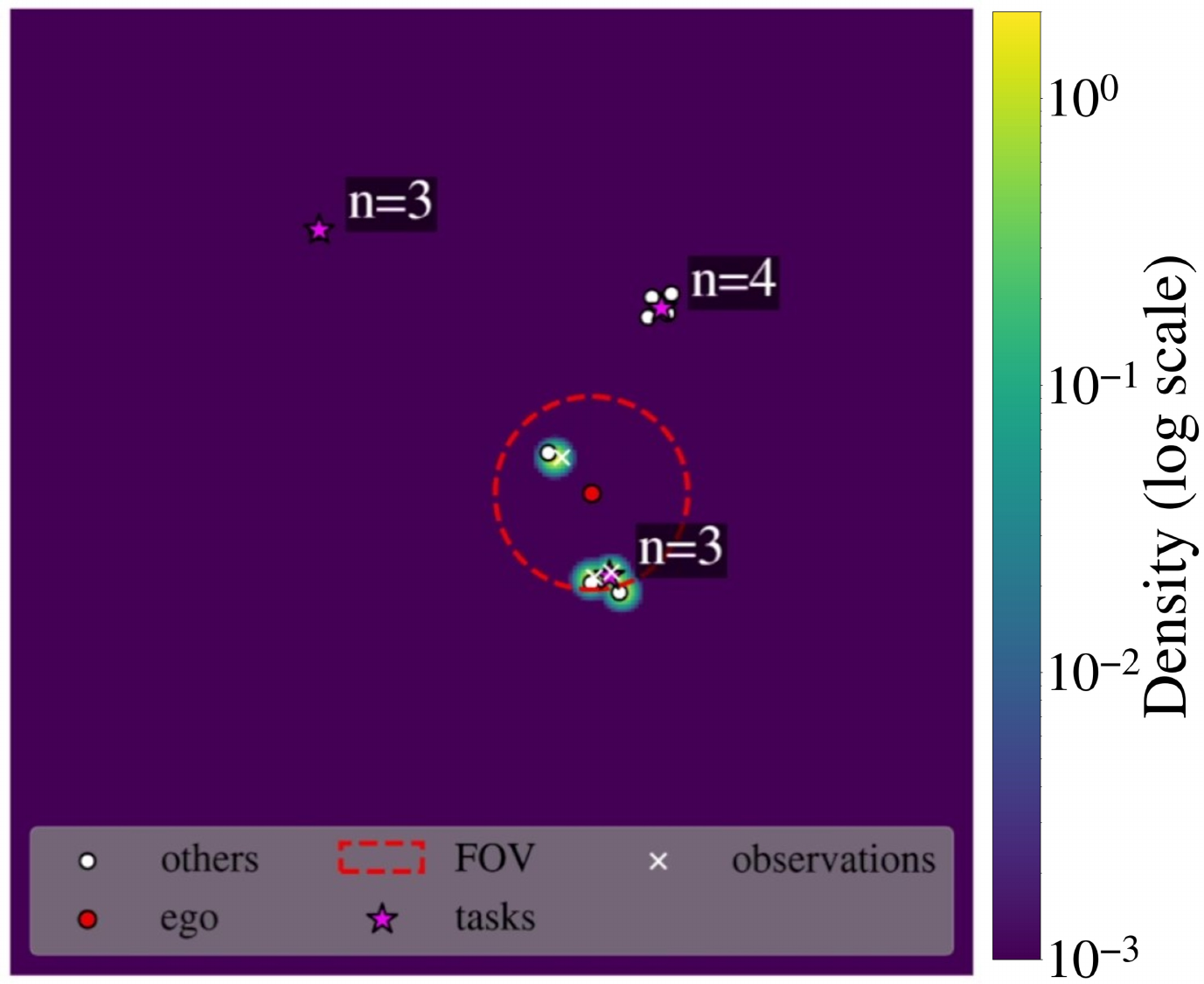}%
    \label{fig:sim_example26}}
    \hfil
    \centering
    \subfloat[]{\includegraphics[height=4.2cm]{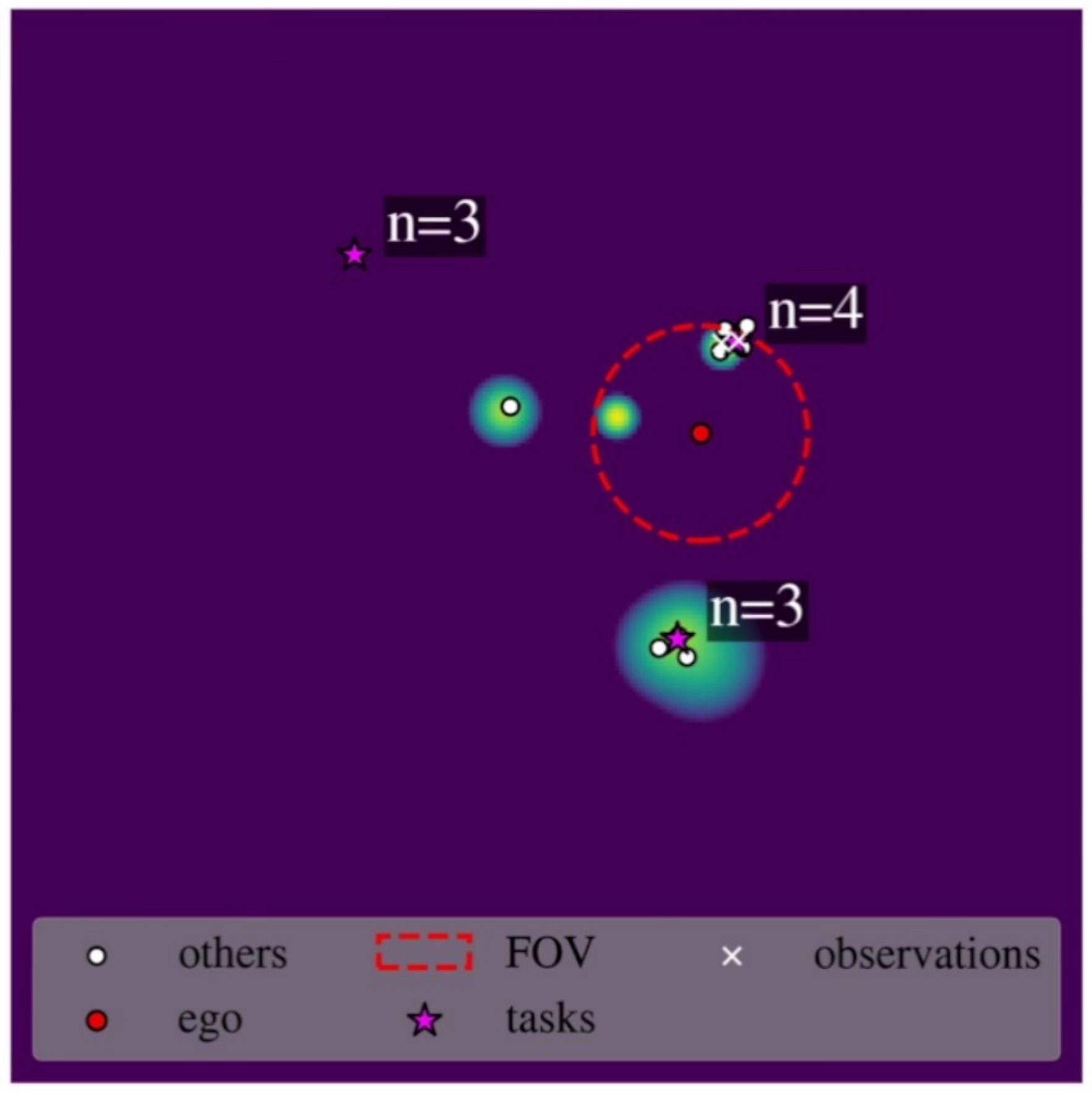}%
    \label{fig:sim_example34}}
    \hfil
    \centering
    \subfloat[]{\includegraphics[height=4.2cm]{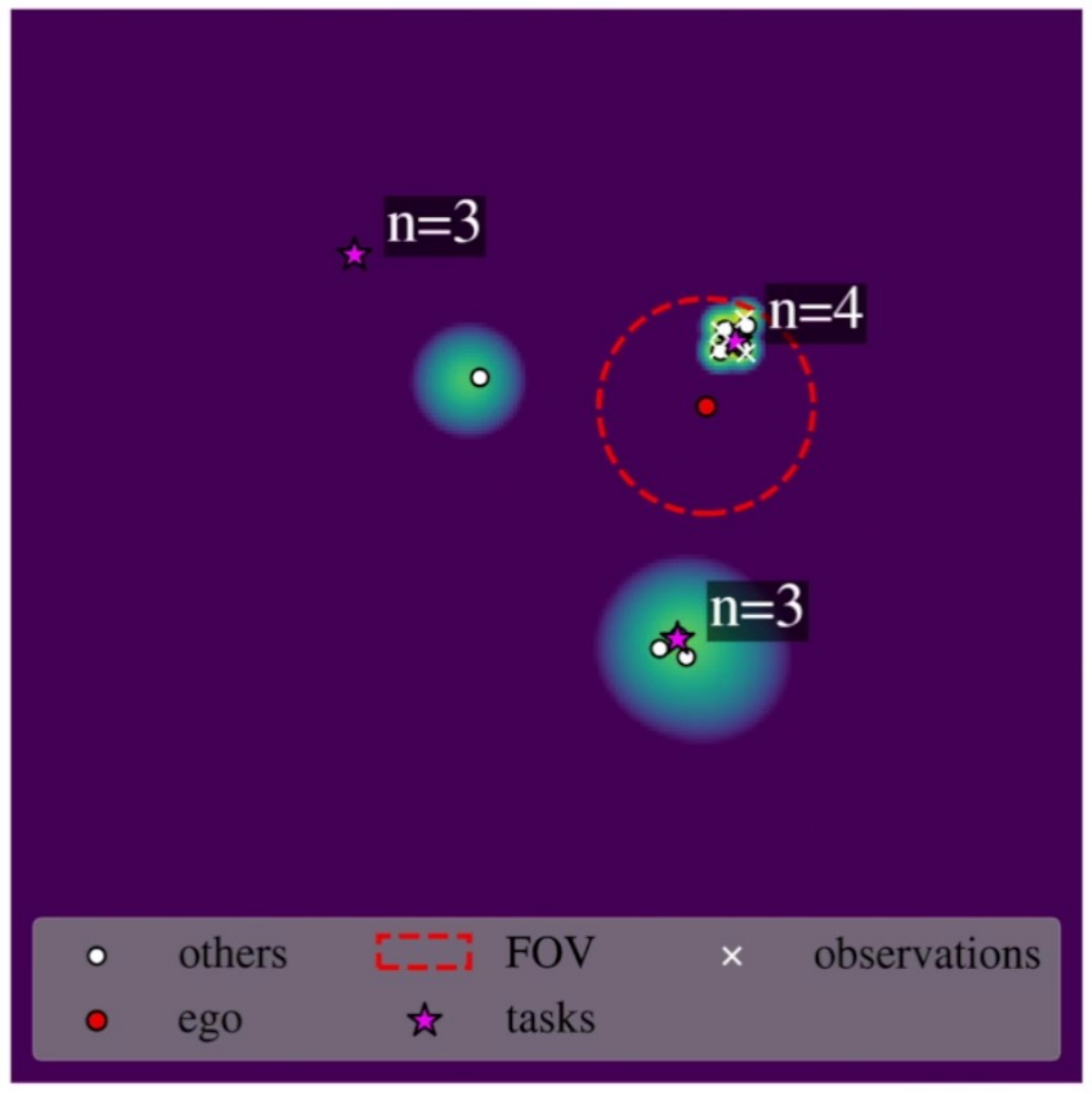}%
    \label{fig:sim_example36}}
    \hfil
    \centering
    \subfloat[]{\includegraphics[height=4.2cm]{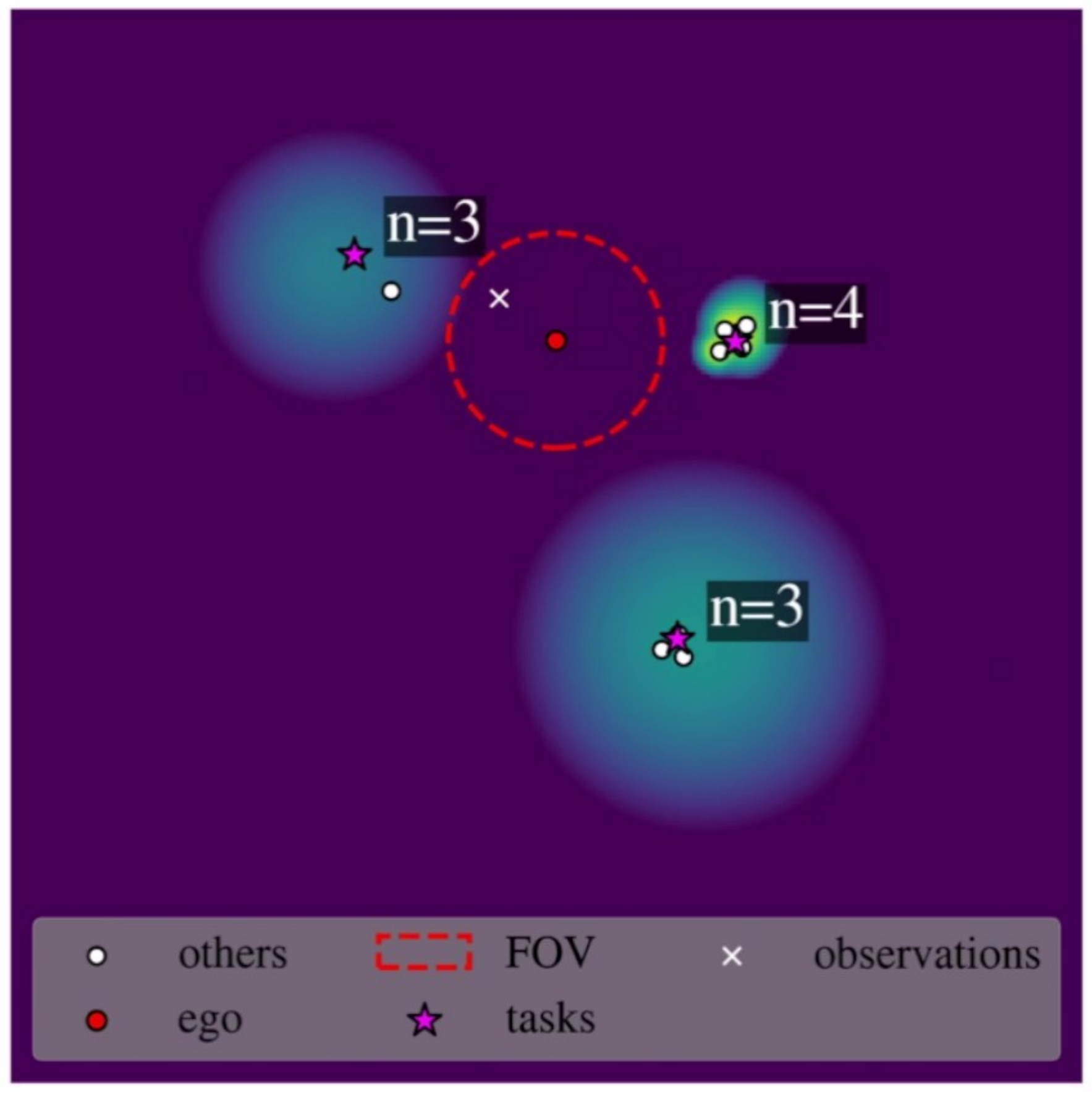}%
    \label{fig:sim_example44}}
    \hfil
    \centering
    \subfloat[]{\includegraphics[height=4.2cm]{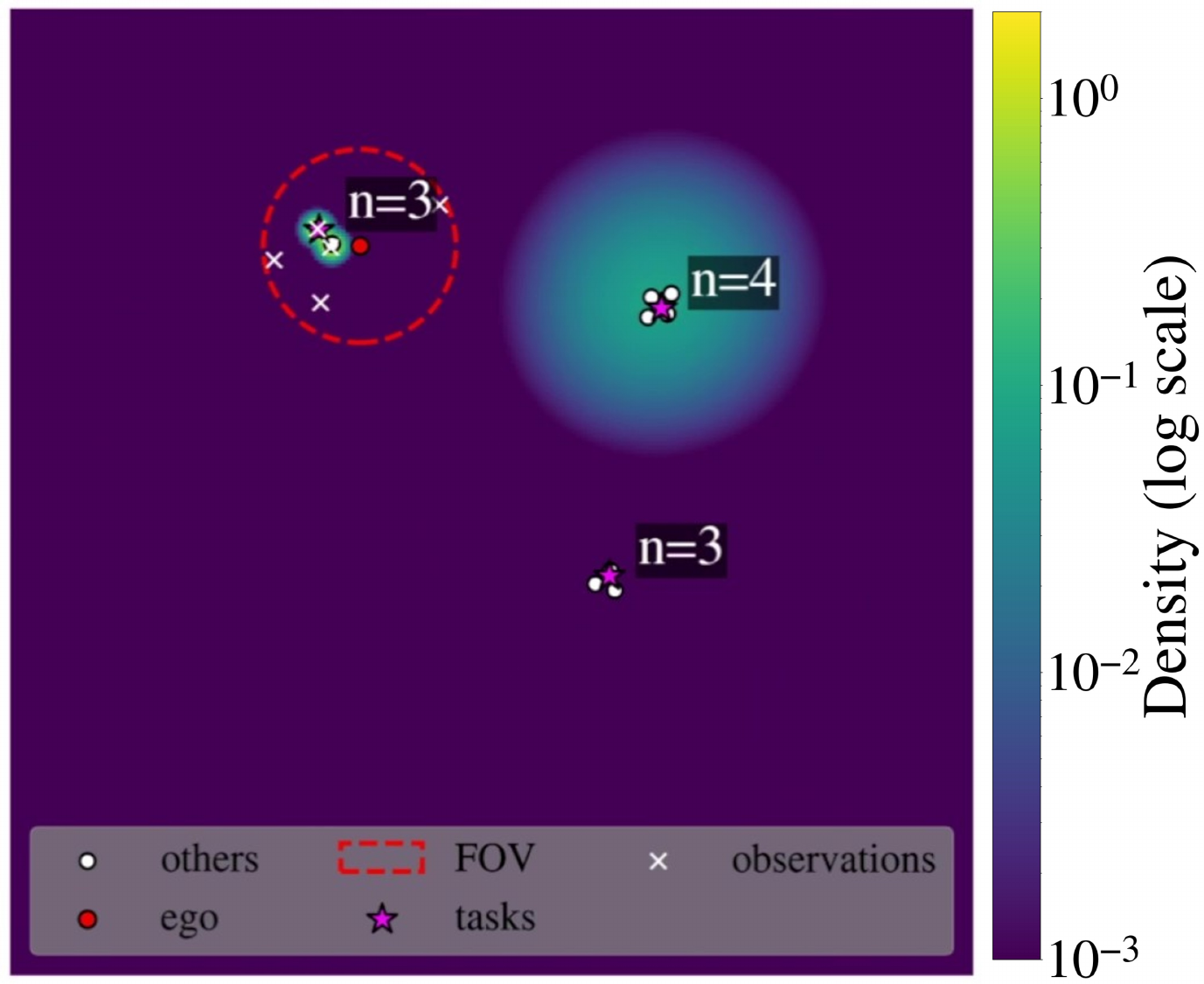}%
    \label{fig:sim_example56}}
    \caption{An example of a randomly generated LMB-based MRTA simulation with 10 robots and 3 tasks. The figure shows how a robot estimates the states of the other robots and continuously updates its task assignment while moving. (a) Initial state. (b)--(h) Snapshots at 8, 18, 26, 34, 36, 44, and 56 s, respectively. All tasks are completed in (h).}
    \label{fig:sim_example}
\end{figure*}

To illustrate the behavior of the proposed LMB-based MRTA framework, Fig.~\ref{fig:sim_example} shows a representative simulation run with a clutter intensity of $\lambda_c=1.0$ in the scenario with $10$ robots and $3$ tasks. Task and robot locations are randomly generated. The red circle denotes the ego robot, white circles denote the other robots, and observation markers (white x) represent both robot detections and clutter.

Initially, the ego robot has no information about other robots and moves toward the nearest task (Fig.~\ref{fig:sim_example}\subref{fig:sim_example0}). As robots enter its sensing range, their states are estimated by the LMB filter, and the task allocation is updated accordingly (Fig.~\ref{fig:sim_example}\subref{fig:sim_example8}).
As additional observations become available, the estimated robot distribution evolves, causing the ego robot to revise its assigned task when other robots occupy its current target task (Fig.~\ref{fig:sim_example}\subref{fig:sim_example18}--\subref{fig:sim_example26}). Although clutter occasionally generates ghost tracks, they are removed as their existence probabilities decrease without supporting observations (Fig.~\ref{fig:sim_example}\subref{fig:sim_example34}--\subref{fig:sim_example36}).
Even after robots leave the field of view, their states continue to be predicted and remain available for task allocation (Fig.~\ref{fig:sim_example}\subref{fig:sim_example36}--\subref{fig:sim_example44}). Eventually, the ego robot converges to a task that can accommodate an additional robot without exceeding the required team size (Fig.~\ref{fig:sim_example}\subref{fig:sim_example56}).

This example demonstrates that the proposed method continuously adapts task assignments under uncertain and intermittent observations without inter-robot communication, while remaining robust to clutter-induced false detections. Although each robot maintains a different estimate of the robot distribution, repeated reassignment gradually reduces inconsistencies as more observations become available.

\begin{figure}[t]
    \centering
    \includegraphics[width=\linewidth]{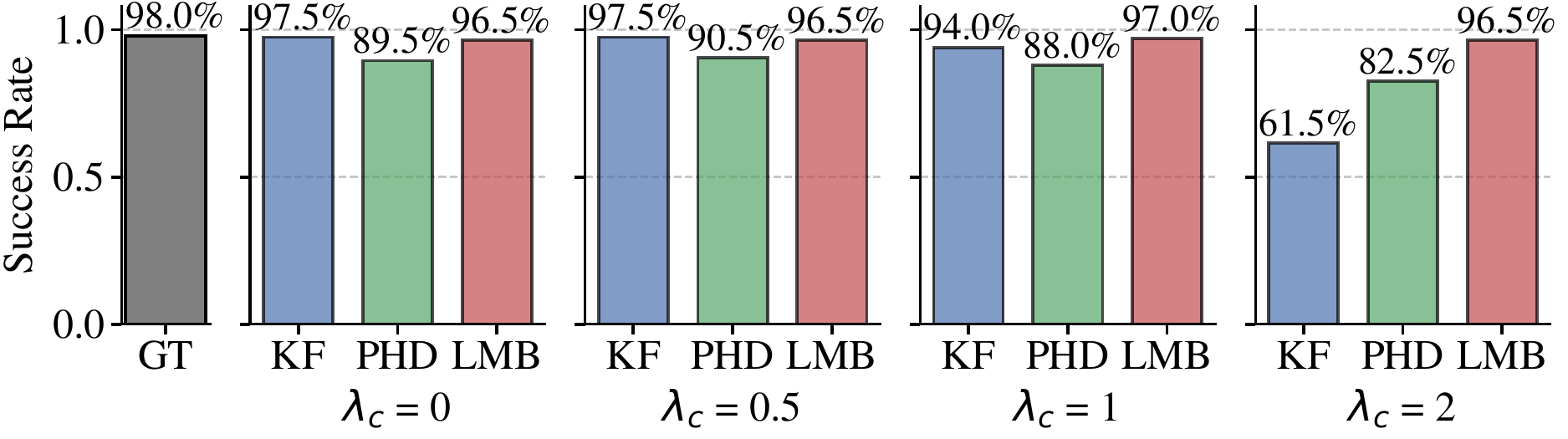}
    \includegraphics[width=\linewidth]{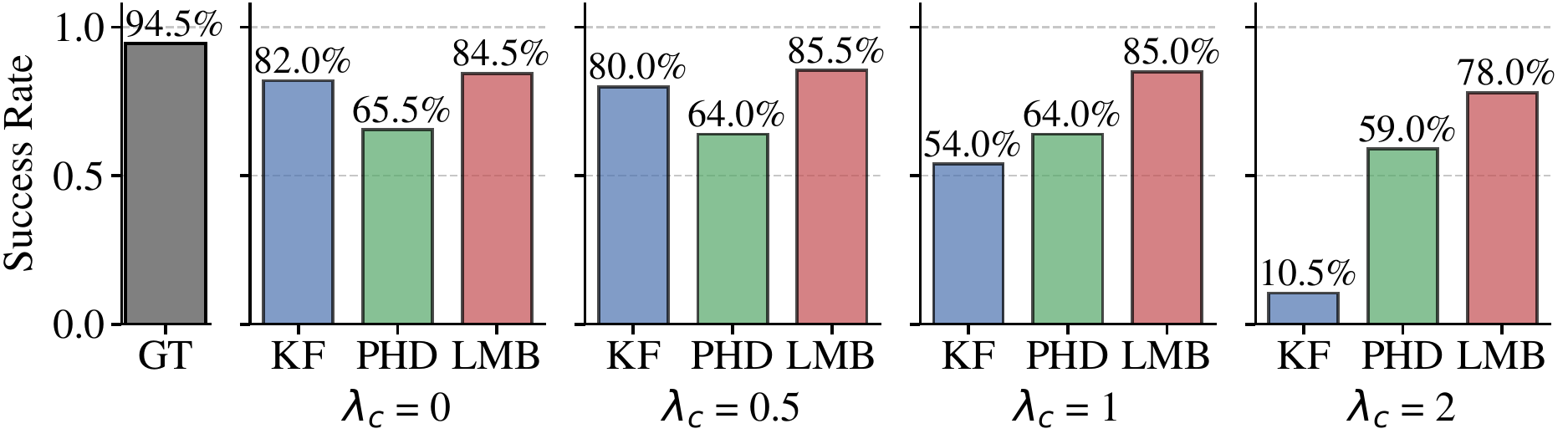}
    \caption{Success rate of task completion under different clutter intensities. The top and bottom panels correspond to the scenarios with 10 robots and 3 tasks, and 15 robots and 4 tasks, respectively.}
    \label{fig:success_rate}
\end{figure}

\begin{figure}[t]
    \centering
    \includegraphics[width=\linewidth]{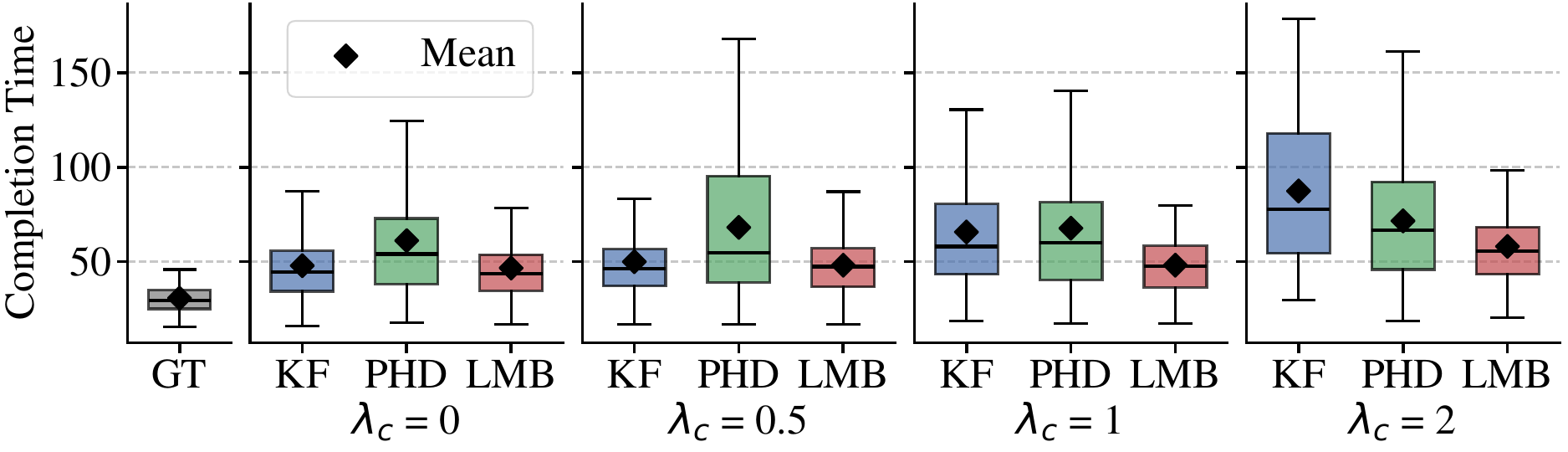}
    \includegraphics[width=\linewidth]{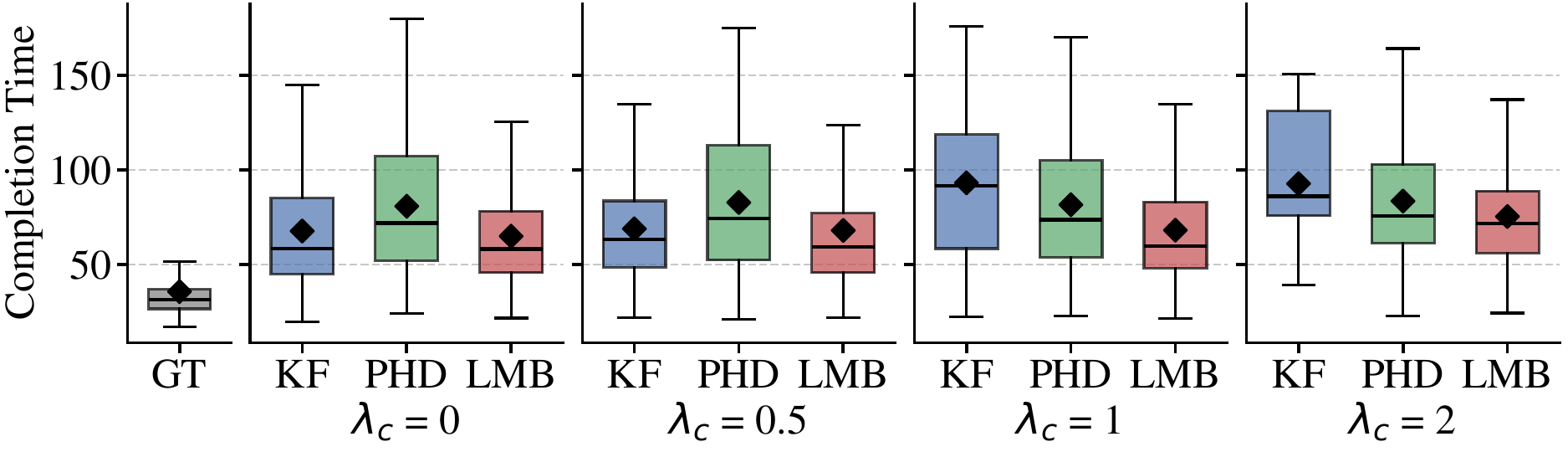}
    \caption{Completion time of successful trials under different clutter intensities. Top: 10 robots and 3 tasks. Bottom: 15 robots and 4 tasks.}
    \label{fig:completion_time}
\end{figure}

\begin{figure}[t]
    \centering
    \includegraphics[width=\linewidth]{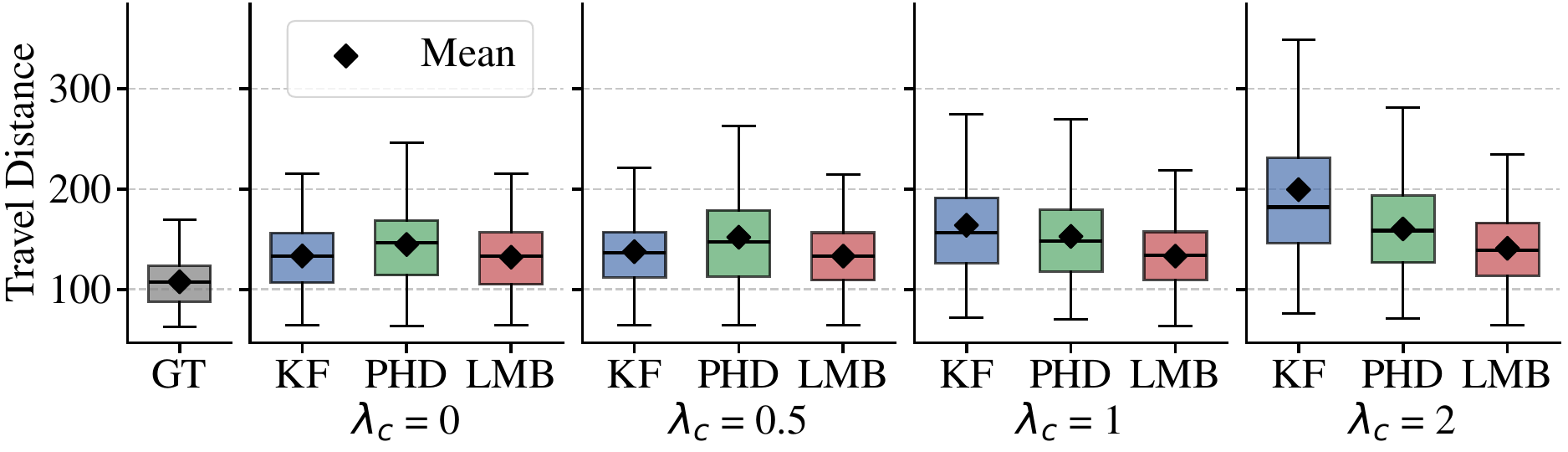}
    \includegraphics[width=\linewidth]{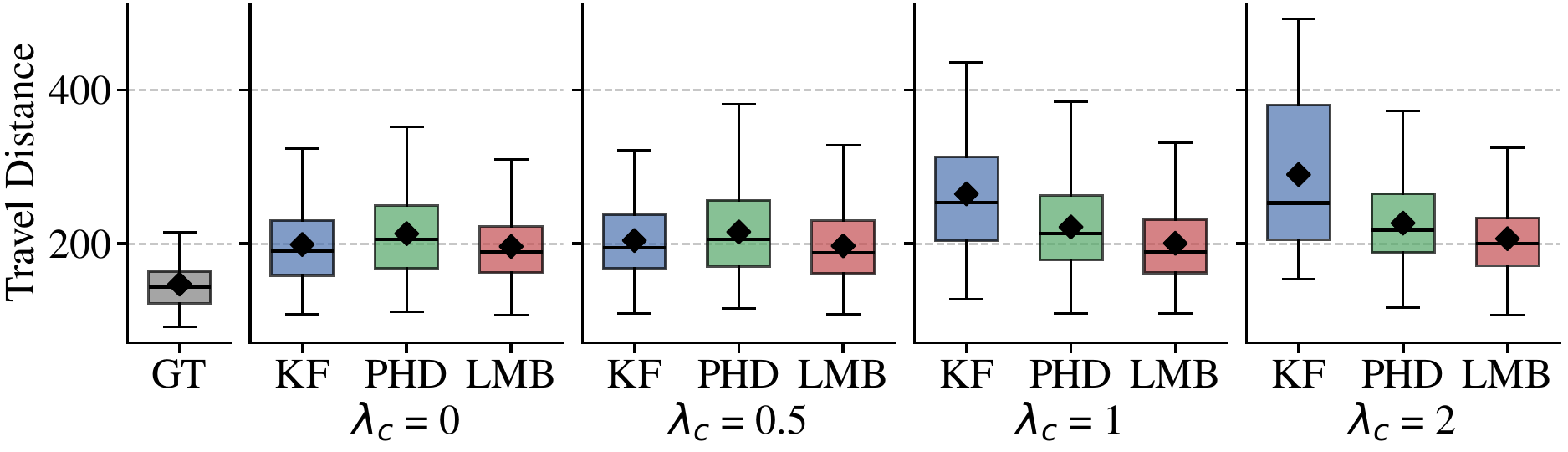}
    \caption{Total travel distance of all robots in successful trials under different clutter intensities. Top: 10 robots and 3 tasks. Bottom: 15 robots and 4 tasks.}
    \label{fig:travel_distance}
\end{figure}

\begin{figure}[t]
    \centering
    \includegraphics[width=\linewidth]{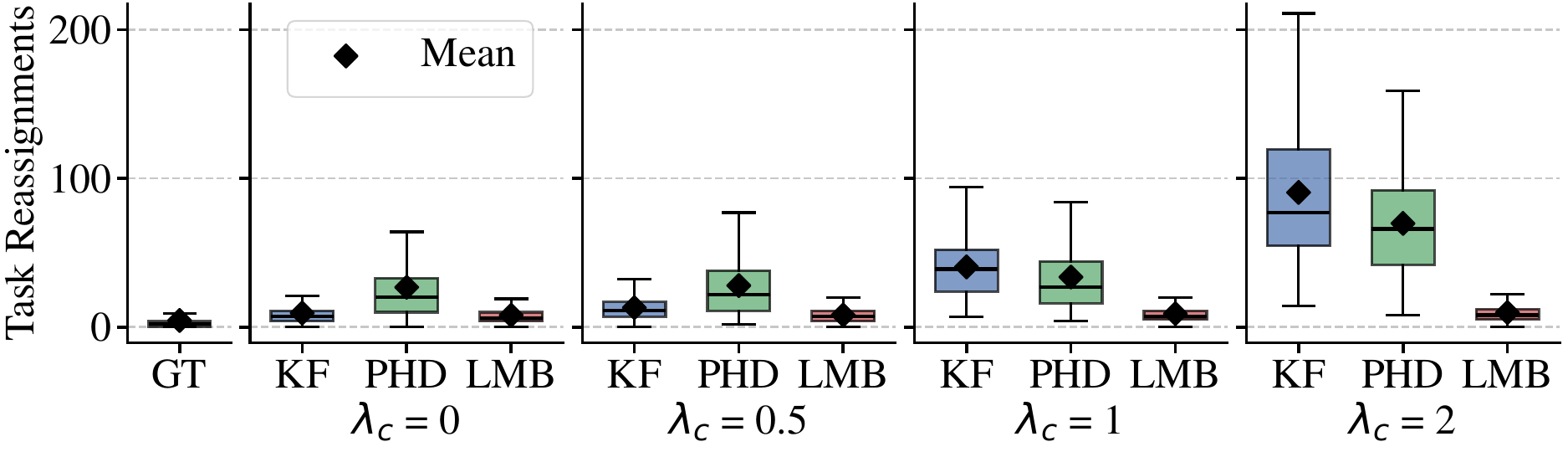}
    \includegraphics[width=\linewidth]{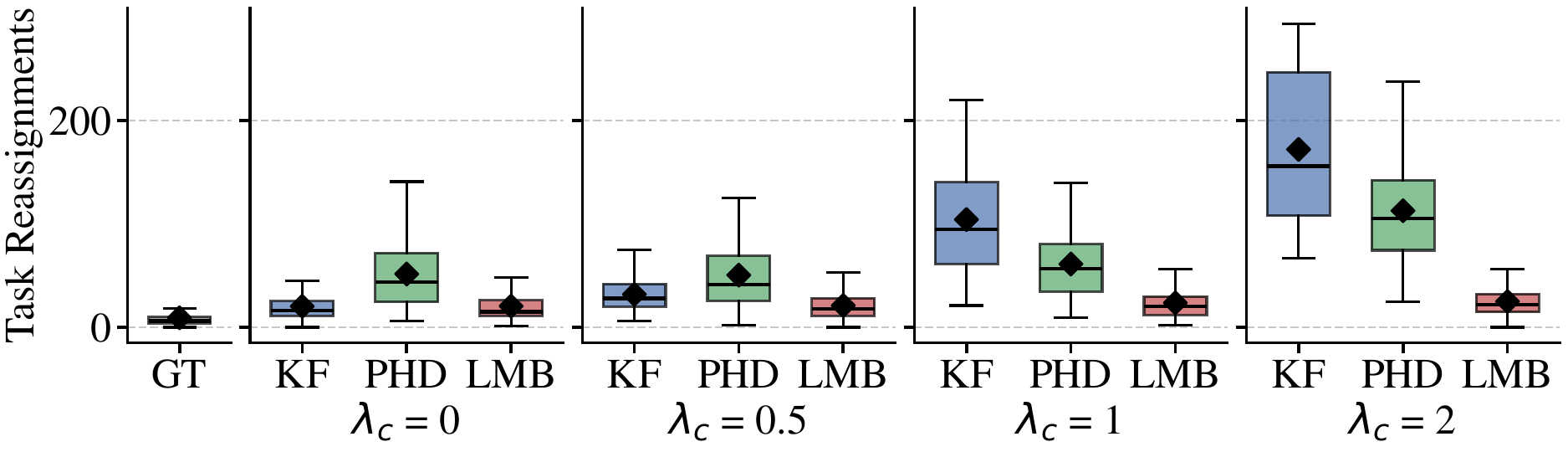}
    \caption{Number of task reassignments in successful trials under different clutter intensities. Top: 10 robots and 3 tasks. Bottom: 15 robots and 4 tasks.}
    \label{fig:task_reassignment}
\end{figure}

Next, we evaluate the proposed framework through Monte Carlo simulations under different clutter intensities and compare it with KF- and PHD-based approaches in the 10-robot/3-task and 15-robot/4-task scenarios. Figures~\ref{fig:success_rate}--\ref{fig:task_reassignment} present the success rate, completion time, travel distance, and task reassignment count. Ground-truth (GT) results are shown on the left of each figure for reference. Note that the success rate of GT is below 100\% because some robots became stuck during the trials due to collision avoidance. Completion time, travel distance, and reassignment count are evaluated only for successful trials and should therefore be interpreted together with the success rate. Overall, performance decreases as the problem size increases; however, both scenarios exhibit similar trends across clutter intensities and estimation methods.

For low clutter ($\lambda_c = 0$ and $0.5$), KF and LMB achieved higher success rates than PHD. Their completion times, travel distances, and reassignment counts were also similar. This is because KF and LMB explicitly maintain individual robot hypotheses, whereas PHD primarily represents the robot distribution through an intensity function and does not directly preserve robot identities. Since the MRTA problem requires satisfying task cardinality requirements \eqref{eq:task_number_satisfaction}, density estimates alone provide limited information for allocation decisions.

At $\lambda_c = 1.0$, PHD showed a lower success rate than KF and LMB in the smaller 10-robot/3-task scenario. In contrast, in the larger 15-robot/4-task scenario, PHD achieved a higher success rate than KF, while other methods exhibited similar completion times, travel distances, and reassignment counts. The LMB-based method maintained almost the same performance as in low-clutter environments, demonstrating robustness against clutter-induced estimation errors.

Under severe clutter ($\lambda_c = 2.0$), the differences became more pronounced. The success rates of KF decreased to 61.5\% and 10.5\% in the 10-robot/3-task and 15-robot/4-task scenarios, respectively, while PHD achieved 82.5\% and 59.0\%. In contrast, the proposed method maintained 96.5\% and 78.0\%, remaining close to its low-clutter performance. Similar trends were observed in completion time and travel distance. The largest difference appeared in task reassignment. Because reassignment is triggered by neighboring-robot estimates, false estimates directly lead to unstable allocation behavior. While KF and PHD showed substantial increases in reassignment counts, LMB consistently maintained a low reassignment count, indicating more reliable allocation decisions.
It should be noted that the success rate of the proposed method also decreased from nearly 100\% in the 10-robot/3-task scenario to approximately 80\% in the 15-robot/4-task scenario, indicating that the performance of the proposed method still degrades as the problem size increases.

Overall, KF performs well in low-clutter environments but is sensitive to false detections. PHD is more robust to clutter, yet its density-based representation is less suitable for MRTA problems with robot cardinality requirements. The proposed LMB-based approach combines robustness to clutter with explicit labeled robot estimates, achieving robust performance across all clutter conditions.